\documentclass[usenames,dvipsnames]{article}
\usepackage{iclr2023_conference,times}

\usepackage{amsmath,amsfonts,bm}

\def\eqref#1{equation~\ref{#1}}

\def\floor#1{\lfloor #1 \rfloor}
\def\1{\bm{1}}

\def\vb{{\bm{b}}}

\def\vd{{\bm{d}}}

\def\vq{{\bm{q}}}

\def\vs{{\bm{s}}}

\def\vu{{\bm{u}}}

\def\mA{{\bm{A}}}

\def\mS{{\bm{S}}}

\def\mW{{\bm{W}}}

\DeclareMathAlphabet{\mathsfit}{\encodingdefault}{\sfdefault}{m}{sl}
\SetMathAlphabet{\mathsfit}{bold}{\encodingdefault}{\sfdefault}{bx}{n}

\newcommand{\R}{\mathbb{R}}

\usepackage{hyperref}
\usepackage{url}

\newcommand{\ours}{\textsc{AligneR}\xspace}

\newcommand{\oursbase}{\textsc{AligneR}$_\text{base}$}
\newcommand{\ourslarge}{\textsc{AligneR}$_\text{large}$}
\newcommand{\oursxl}{\textsc{AligneR}$_\text{xl}$}
\newcommand{\oursxxl}{\textsc{AligneR}$_\text{xxl}$}

\newcommand{\msmarco}{MS MARCO}
\newcommand{\nfcorpus}{NFCorpus}
\newcommand{\bioasq}{BioASQ}
\newcommand{\nq}{NQ}
\newcommand{\hotpotqa}{HotpotQA}
\newcommand{\fiqa}{FiQA}

\newcommand{\treccovid}{TREC-COVID}

\newcommand{\arguana}{ArguAna}
\newcommand{\touche}{Touché-2020\xspace}
\newcommand{\quora}{Quora}
\newcommand{\dbpedia}{DBPedia}
\newcommand{\scidocs}{SCIDOCS}
\newcommand{\fever}{FEVER}
\newcommand{\climatefever}{Climate-FEVER}
\newcommand{\scifact}{SciFact}

\usepackage{color,soul}
\setulcolor{white}
\newcommand{\hlr}[2]{\setlength{\fboxsep}{0.2pt}\colorbox{Goldenrod!#2}{\rule[-.05\baselineskip]{0pt}{.7\baselineskip}{\ul{#1}}}}

\usepackage{array}
\usepackage{amsmath}
\usepackage{amsfonts}
\usepackage{graphicx}
\usepackage{bm}
\usepackage{bbm}
\usepackage{booktabs}
\usepackage{xcolor} %
\usepackage{subcaption}
\usepackage{booktabs} %
\usepackage{tikz} %
\usepackage{tikz-layers} %
\usepackage{enumitem}
\usepackage{footnote}
\usepackage{mathtools}
\usepackage{multicol}
\usepackage{latexsym}
\usepackage{mathptmx}
\usepackage{multirow}
\usepackage{natbib}
\usepackage{outlines}
\usepackage{pifont}%
\usepackage{soul}
\usepackage{rotating}
\usepackage{tabulary}
\usepackage{arydshln}
\usepackage{floatrow}
\usepackage[flushleft]{threeparttable}
\usepackage{blindtext}

\usepackage{caption}
\usepackage{url}
\usepackage{xspace}
\usepackage[utf8]{inputenc}
\usepackage{hyperref}
\usepackage{cleveref}
\usepackage{diagbox}
\usepackage{tablefootnote}
\usepackage{pgfplots}
\usepackage{wrapfig}

\newcommand{\cmark}{\ding{51}}%
\newfloatcommand{capbtabbox}{table}[][\FBwidth]
\Crefformat{section}{\S#2#1#3}

\title{Multi-Vector Retrieval as Sparse Alignment}

\iclrfinalcopy

\author{Yujie Qian\thanks{Equal contribution.} \thanks{Currently at MIT CSAIL. Work done during Google internship.} , Jinhyuk Lee, Sai Meher Karthik Duddu, Zhuyun Dai, Siddhartha Brahma, \\ \textbf{Iftekhar Naim, Tao Lei, Vincent Y.~Zhao$^{*}$\thanks{Correspondence: \url{vzhao@google.com}}}\\
Google Research
}

\begin{document}

\maketitle

\begin{abstract}

Multi-vector retrieval models improve over single-vector dual encoders on many information retrieval tasks.
In this paper, we cast the multi-vector retrieval problem as \textit{sparse alignment} between query and document tokens. 
We propose \ours, a novel multi-vector retrieval model that learns sparsified pairwise alignments between query and document tokens (e.g. `\textit{dog}' vs. `\textit{puppy}') and per-token unary saliences reflecting their relative importance for retrieval.
We show that controlling the sparsity of pairwise token alignments often brings significant performance gains.
While most factoid questions focusing on a specific part of a document require a smaller number of alignments, others requiring a broader understanding of a document favor a larger number of alignments. 
Unary saliences, on the other hand, decide whether a token ever needs to be aligned with others for retrieval (e.g. `\textit{kind}' from `\textit{what \underline{kind} of currency is used in new zealand}').
With sparsified unary saliences, we are able to prune a large number of query and document token vectors and improve the efficiency of multi-vector retrieval.
We learn the sparse unary saliences with entropy-regularized linear programming, which outperforms other methods to achieve sparsity.
In a zero-shot setting, \ours scores 51.1 points nDCG@10, achieving a new retriever-only state-of-the-art on 13 tasks in the BEIR benchmark. In addition, 
adapting pairwise alignments with a few examples ($\leq$ 8) further improves the performance up to 15.7 points nDCG@10 for argument retrieval tasks.
The unary saliences of \ours helps us to keep only 20\% of the document token representations with minimal performance loss.
We further show that our model often produces interpretable alignments and significantly improves its performance when initialized from larger language models.

\end{abstract}

\section{Introduction}

Neural information retrieval (IR) has become a promising research direction for improving traditional IR systems.
The most-commonly adopted approach called the dual encoder operates by representing every query and document as a single dense vector.
Given sufficient annotations, dual encoders directly learn task-driven similarity between vectors, and often surpass traditional IR systems on complex tasks such as question answering~\citep{lee2019latent,dpr,gtr}.
However, these models can struggle to generalize over out-of-domain datasets~\citep{beir} and/or entity-centric questions~\citep{sciavolino2021simple} due to the limited representational capacity of single vectors.
As a remedy, multi-vector retrieval models~\citep{ colbert,luan2021sparse,coil} instead use multiple vectors, typically the contextualized token vectors, to represent the text. 
These models largely improve the model expressiveness, and exhibit much stronger performance and robustness compared to their single-vector counterparts.

Existing multi-vector retrieval models such as ColBERT~\citep{colbert} computes query-document similarity by selecting the highest scoring document token for each query token and aggregating the scores.
This sum-of-max method has two major limitations.
First, restricting the selection to a \emph{single} document token can be highly sub-optimal for some retrieval tasks.
As we will show in our experiments, the retrieval performance can be improved by more than 16 points nDCG@10 by relaxing this constraint.
Second, the method also leads to a large search index and expensive computation.
Specifically, the retrieval and storage cost scales linearly with the query and document length, making multi-vector retrieval models an inferior choice for efficiency-demanding applications.
We directly tackle these challenges to build faster and more accurate models.

\begin{figure}
    \centering
    \includegraphics[width=0.98\linewidth]{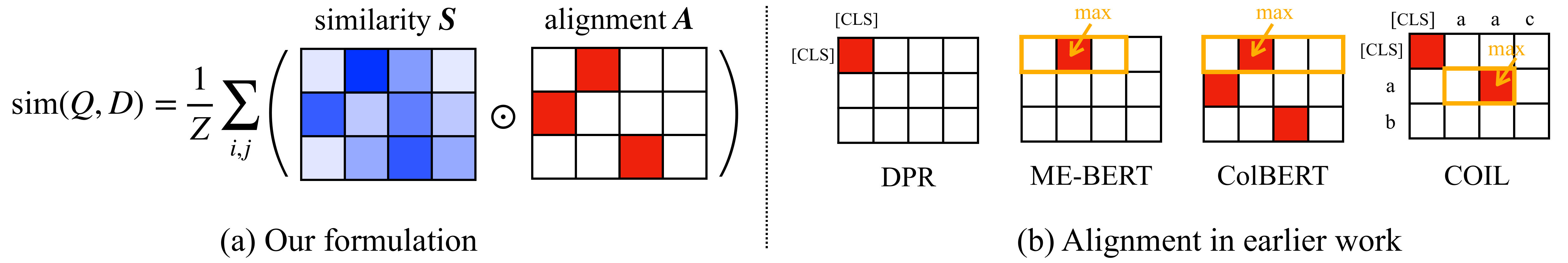}
    \caption{
    (a) We formulate multi-vector retrieval as token-level sparse alignment; (b) Earlier models can be covered by our formulation as using different alignments.
    }
    \label{fig:model}
\end{figure}

The representation learning problem of multi-vector retrieval can be formulated as optimizing token-level alignment.
Specifically, we use a \textit{sparse alignment matrix} to aggregate token-level similarities, where each element indicates the alignment of a pair of tokens.
From this point of view, we are able to formulate different retrieval models in a unified manner~(\Cref{fig:model}) and discern the drawbacks of existing models.

Based on our formulation, we propose \ours, a novel multi-vector retrieval model that consists of pairwise alignment and unary salience.
Pairwise alignments form the basis of \ours, where pairs of query and document tokens are sparsely aligned based on their contextual representations. It is discovered that changing the sparsity of alignment can significantly impact the performance on retrieval tasks.
For instance, factoid questions often favor a small number of alignments since they often focus on a small part of a document. However, other queries for different tasks (e.g., argument retrieval and fact checking) require a larger number of alignments for a broader understanding of a document.
Our findings also support the claim of~\citet{promptagator} that retrieval tasks with different intents should be modeled differently.

\ours also learns unary saliences, which decides whether each token ever needs to be aligned with any other token for retrieval.
This corresponds to masking an entire row or column of the alignment matrix, rather than individual token alignments.
To sparsify entire rows or columns, we introduce an algorithm that produces sparse token salience and is end-to-end differentiable based on a novel formulation of entropy-regularized linear programming.
Sparsified unary saliences allow us to prune a large number of document and query token representations, making multi-vector retrieval a more efficient and affordable solution.

We evaluate \ours on the BEIR benchmark~\citep{beir}, which covers a diverse set of retrieval tasks in multiple domains.\footnote{We will release our model checkpoints to encourage future research.}
In a zero-shot setting, we show that simply scaling our model achieves the state-of-the-art performance, outperforming prior neural retrievers without contrastive pre-training, model-based hard negative mining, or distillation.
By adapting the pairwise alignments with a few examples from the target task --- similar to the setup of \citet{promptagator} --- \text{\ours} can be further improved by up to 15.7 points nDCG@10 on argument retrieval tasks.
Meanwhile, pruning with our unary saliences can reduce 50\% of query tokens for better run-time efficiency and 80\% of document tokens for better storage footprint, with less than 1 point decrease of nDCG@10.
The pairwise alignments and unary saliences are also highly interpretable so that they often serve as concise rationales for retrieval.

\section{Multi-Vector Retrieval as Sparse Alignment}

Given a query $Q$ and a collection of $N$ documents $\mathcal{C}=\{D^{(1)}, \dots, D^{(N)}\}$, a key problem in retrieval is how to represent these textual inputs in order to facilitate efficient search.
To this end, one approach is lexical retrieval using sparse bag-of-words representation of the text; the other approach is dense retrieval, which this work focuses on. Dense retrieval models learn a parameterized function that encodes the query and documents into query representation $\vq$ and document representations $\{\vd^{(1)}, \dots, \vd^{(N)}\}$ respectively. 
Typically, each representation is a single $d$-dimensional vector.
For retrieval, the similarity function is often defined as $\text{sim}(Q, D^{(i)}) = \vq^\top \vd^{(i)}$, and documents having high similarity scores to the query are retrieved.

\subsection{Multi-vector Retrieval}
Instead of representing each query and document as a single fixed-length vector, multi-vector retrieval represents them with multiple token vectors, mainly to improve the limited capacity of fixed-length representations.
Specifically, a query $Q=\{q_1, \dots, q_n\}$ and a document $D=\{d_1, \dots, d_m\}$ are encoded into a set of vectors $\{\vq_1, \dots, \vq_n\}$ and $\{\vd_1, \dots, \vd_m\}$.
The similarity function between a query and a document is re-defined for multi-vector retrieval.
For instance, ColBERT~\citep{colbert} designs the similarity function as follows:
\begin{equation*}
    \text{sim}(Q,D) = \sum_{i=1}^n \max_{j=1\dots m} \vq_i^\top \vd_j.
\end{equation*}
For retrieval, instead of indexing $N$ document vectors, multi-vector retrieval pre-computes $N \times \bar{m}$ document \textit{token} vectors where $\bar{m}$ is the average length of documents.
Then, it retrieves $K$ document token vectors for each query token vector with Maximum Inner-Product Search (MIPS), resulting in $n \times K$ candidate document tokens.
The retrieved tokens are used to trace back the original documents~\citep{lee2021phrase}, often followed by a final refinement stage that scores the similarity $\text{sim}(Q, D)$ with all token representations of each document and the query~\citep{colbert}.
We adopt the same practice of ColBERT in our experiments.

\subsection{Sparse Alignment Formulation}
\label{sec:formulation}
A key design question for retrieval models is defining the similarity function in a manner that balances model expressiveness and inference cost.
To facilitate our discussion, we formalize the similarities used in previous methods into a class of sparse alignment functions.
The formulation also leads to a principled extension over existing work, which we will describe in \Cref{sec:model}.

We begin by defining a \textit{similarity matrix} $\mS\in\R^{n\times m}$ computed from all pairs of query and document tokens, where $\mS_{i,j}=\vq_i^\top \vd_j$.
Then, we use an \textit{alignment matrix} $\mA\in[0,1]^{n\times m}$ to compute the similarity between $Q$ and $D$ as follows:
\begin{equation}
    \text{sim}(Q,D) = \frac{1}{Z}\sum_{i=1}^n \sum_{j=1}^m \mS_{i,j}\mA_{i,j} 
    \label{eq:sim}
\end{equation}
where $Z$ is a normalization term defined as $Z=\sum_{i,j}\mA_{i,j}$. 
The alignment matrix $\mA$ can be directly derived from $\mS$ or computed as a function of $Q$ and $D$.

On the top of our formulation, the alignment matrix $\mA$ is constrained to be \textit{sparsely} activated: $||\mA||_0 \leq \sigma$ where $||\cdot||_0$ is the number of non-zero elements in a matrix.
Sparse activation assumes that only a few query-document token matches are critical for retrieval, inspired by traditional retrieval methods.
Indeed, most existing dense retrieval models already enforce the sparse alignment with their own heuristics.
\Cref{fig:model} illustrates how different models can be described under our formulation:

\begin{itemize}
\item \textbf{Dense passage retriever} (DPR; \citealp{dpr}) uses a single [CLS] vector to represent each query and document.
This is equivalent to setting $A_{1,1}=1$ and $0$ otherwise, resulting in $||\mA||_0 = 1$.
\item \textbf{ME-BERT}~\citep{luan2021sparse} uses the first $k$ document token vectors for multi-vector representations of documents but a single vector for query. The similarity function is $\max_{j=1\dots k} \vq_1^\top\vd_j$, which is equivalent to setting $A_{1,j}=1$ when $\mS_{1,j}$ is the maximum within $\mS_{1,1}$ to $\mS_{1,k}$, and 0 otherwise. The alignment sparsity is $||\mA||_0=1$.
\item \textbf{ColBERT} uses the sum-of-max similarity function $\sum_{i=1}^n\max_{j=1\dots m}\mS_{i,j}$ that is equivalent to setting an alignment matrix to select the maximum element from each row of $\mS$, i.e., $\mA_{i,j}=1$ when $\mS_{i,j}$ is the maximum within $\mS_{i,:}$.
$||\mA||_0=n$ in this case.
\item 
\textbf{COIL}~\citep{coil}, similar to ColBERT, also selects the maximum element from each row of $\mS$, but requires a lexical exact match for a selected pair, i.e., $\mA_{i,j}=1$ when $\mS_{i,j}$ is the maximum within $\{\mS_{i,j'} \mid  q_i = d_{j'} \}$.
$||\mA||_0 \leq n$ in this case. 
\end{itemize}

The choice of similarity and sparsity can have a large impact on model capacity and efficiency. 
For instance, ColBERT is more expressive and robust than DPR~\citep{beir}, but its retrieval and storage costs are much higher.
Our work seeks to further advance expressiveness while retaining a strong efficiency.
We describe our method in the next section.

\section{\ours}
\label{sec:model}

\begin{figure}
    \centering
    \includegraphics[width=0.98\linewidth]{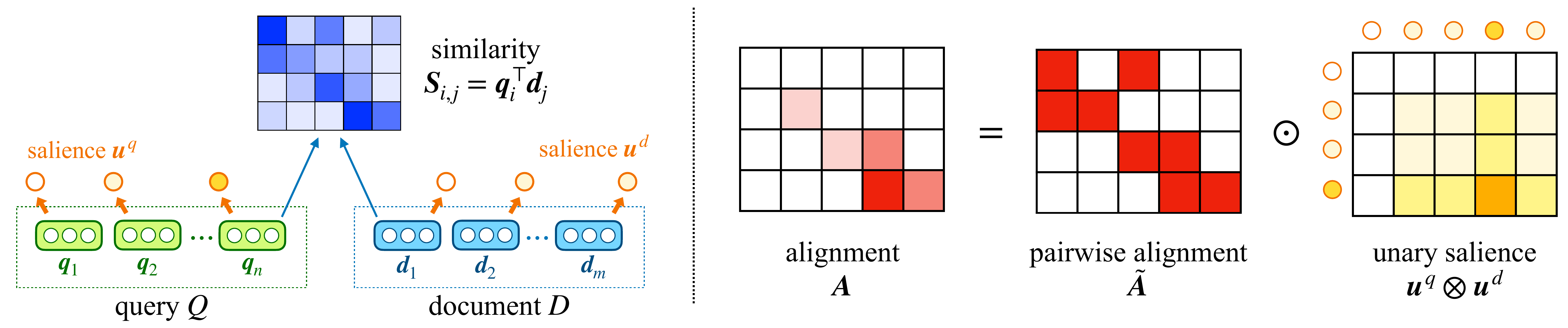}
    \caption{\ours factorizes the alignment matrix into pairwise alignments and unary saliences. Pairwise alignment focuses on the alignment of individual token pairs. Unary saliences are determined by per-token salience features. 
    }
    \label{fig:aligner}
\end{figure}

In this section, we present \ours built upon the sparse alignment formulation.
\ours factorizes the alignment matrix into pairwise alignment and unary salience:
\begin{equation}
    \mA = \tilde{\mA} \odot (\vu^q \otimes {\vu^d})
    \label{eq:factorize}
\end{equation}
where $\odot$ is the Hadamard product and $\otimes$ is the outer product of two vectors.
Pairwise alignment $\tilde{\mA}\in\R^{n\times m}$ determines which pairs of query and document tokens should be aligned, with the sparsity constraints tailored for downstream tasks (\Cref{sec:adaptation}). 
Unary salience $\vu^q\in\R^n$ and $\vu^d\in\R^m$ are sparse token weights deciding whether a token ever needs to be aligned (\Cref{sec:coarse}).

The factorization is introduced based on two critical hypotheses.
First, the optimal sparsity of alignment can be \emph{task-dependent}. Instead of imposing top-1 constraint as in ColBERT, activating more than one alignments for a query token can enhance retrieval performance for certain tasks.
In our analyses for instance, we observe factoid questions that only concern a specific part of a document require a small number of alignments, while some other queries (such as fact checking) require more alignments for a broader understanding of the document.
We explore different search spaces of the pairwise alignment matrix $\tilde{\mA}$ in order to achieve better retrieval performance for each downstream task.
Second, alignment is only needed for \emph{very few} tokens.
For example, we analyzed 2000 most retrieved documents in our preliminary study, and found only 12.8\% document tokens are retrieved by at least one query.\footnote{The analysis was performed on MS MARCO~\citep{msmarco} using our implementation of ColBERT.}
Intuitively, tokens that are uninformative do not need to be aligned and stored, corresponding to sparse activation over an entire row or column of $\mA$.
\ours directly learns the row and column sparsity as unary salience, and utilizes them to enhance retrieval efficiency.

\subsection{Adapting Pairwise Alignment}\label{sec:adaptation}
Queries and documents can have varied distributions. For example, a query can be a single entity, a natural question, or a few sentences, and a document can range from a short paragraph to a long article. The search intent also changes from task to task~\citep{promptagator}. These changes can lead to different optimal alignment strategies.
We explore the following sparse alignment variants that go beyond the top-1 strategy commonly adopted in existing work:
\begin{itemize}
    \item 
\textbf{Top-$k$}.
Each query token is aligned with $k$ document tokens with highest similarity scores. Precisely, $\tilde{\mA}_{i,j}=1$ when the $j$-th token is within top-$k$ of the row $\mS_i$. When $k=1$, it is equivalent to ColBERT.
    \item
\textbf{Top-$p$}.
This strategy is similar to top-$k$, but instead of aligning each query token with exactly $k$ tokens, it makes the number of alignments proportional to the document length, i.e., each query token aligns with $\max(\floor{p\cdot m}, 1)$ tokens where $m$ is the document length and $p \in [0, 1]$ is the alignment ratio. 
\end{itemize}

Despite their simplicity, these variants can indeed enhance retrieval accuracy significantly on tasks such as argument retrieval.
More importantly, while it is possible to train separate models for different alignment variants, we are interested in fast test-time adaptation using a single shared model as many important retrieval tasks lack sufficient training data~\citep{beir}.
Specifically, we first train \ours using a fixed alignment strategy such as top-1 in a source domain, and adapt the alignment strategy to each target task without changing the model parameters.\footnote{We have also explored a differentiable alignment with sparsity constraints (\Cref{sec:diffalign}), but alignment adaptation is still necessary to achieve good performance on target tasks.}
We use the following few-shot alignment adaptation method.
Given a corpus $\{D^{(1)},\dots,D^{(N)}\}$, and a few relevance-annotated query-document pairs from the target task $\{(Q^1, D^1_+),\dots\, (Q^K, D^K_+)\}$, we first retrieve candidate documents with the learned token representations, and decide the pairwise alignment strategy based on the ranking performance on the annotated data. 
This adaptation can be performed efficiently because the alignment only concerns the computation of similarity score (Eq.~\ref{eq:sim}) in the refinement stage. 
In practice, for some tasks,  we are able to find a well-suited alignment strategy and improve the retrieval performance with as few as 8 annotated examples.

\subsection{Learning Unary Salience}
\label{sec:coarse}

\ours predicts token saliences from their token representations. For brevity, we only present the formulation for document salience, and query salience is defined similarly. Specifically, the salience of the $i$-th document token $u^d_i$ is defined as:
\begin{equation}
    u^d_i=\lambda_i^d \cdot f(\mW^d\vd_i+b^d)
\label{eq:salience}
\end{equation}
where $\mW^d$ and $b^d$ are learnable parameters. $f$ is a non-linear activation function and we use ReLU such that salience is always non-negative. $\bm{\lambda}^d=\{\lambda^d_i\}$ are gating variables to control the overall sparsity of $\vu^d$, which we will elaborate next. 

For the document salience to be meaningful, we  enforce salience sparsity as an inductive bias.
\ours jointly optimizes sparse salience with other parts of the model.
Since tokens with zero salience do not contribute to computing similarity,
our model will be encouraged to identify more important tokens in order to retain good retrieval performance.
Note that during training we do not have any explicit annotation on which tokens are important.
Instead, $\vu^d$ (and similarly $\vu^q$) are directly optimized to minimize the training loss, under the sparsity constraint that $\|\bm{\lambda}^d\|_0 = \left\lceil\alpha^d \cdot m\right\rceil$, where $\alpha^d$ is a constant sparsity ratio and $m$ is the document length.

Of course, a key question is how we can optimize the unary salience component given the controlled sparsity.
We leverage a novel technique called entropy-regularized linear programming to enable end-to-end optimization.
Specifically, let $k=\left\lceil \alpha^d\cdot m\right\rceil$ denotes the desired sparsity, $s_i = f(\mW^d\vd_i+b^d)$ denotes the token score before the sparse gate $\lambda^d_i$ is applied, and $\vs, \bm{\lambda}^d \in\R^m$ be the vectors $\{s_i\}$ and $\{\lambda^d_i\}$ respectively. 
$\bm{\lambda}^d$ is computed by solving the following optimization problem:
\begin{equation}
    \max_{\bm{\lambda}} \ \vs^\top \bm{\lambda} + \epsilon H(\bm{\lambda})
    \qquad \text{s.t. } \ \ \bm{1}^\top\bm{\lambda} = k, \ \  \lambda_i\in [0, 1],\ \forall i=1,\dots,m.
\label{eq:opt}
\end{equation}
where $H(\cdot)$ is the elementwise entropy function\footnote{$H(\bm{\lambda}) = \sum_{i=1}^m -\lambda_i \log\lambda_i$} and $\epsilon > 0$ is a small constant.
The optimization can be seen as a relaxed top-$k$ operation. Without the entropy term $\epsilon H(\cdot)$, it becomes an instance of linear programming where the solution $\bm{\lambda}^d$ is a binary mask indicating the top-$k$ values of $\vs$, i.e., $\lambda^d_i = 1$ if and only if $s_i$ is one of top-$k$ values in $\vs$.
This top-$k$ optimization is smoothed by adding the small entropy term $\epsilon H(\cdot)$ and by relaxing $\lambda_i$ from exact binary to $[0, 1]$.
Given small $\epsilon$, this still produce a sparse solution $\bm{\lambda}^d$ and can be solved using simple vector operations.
Specifically, let $a \in\R$ and $b_i\in\R$ for $i=1,\cdots ,m$ be auxiliary variables that are initialized to zero. 
We iteratively update these variables using the following equations:
\begin{equation}
    a' = \epsilon \ln(k) - \epsilon \ln\left\lbrace\sum_i \exp\left(\frac{s_i + b_i}{\epsilon}\right)\right\rbrace, \qquad b'_i = \min(-s_i - a', 0).
\label{eq:iterations}
\end{equation}
In practice, it is sufficient to run only a few iterations and the final solution is given by 
$\lambda_i = \exp(\frac{s_i+b_i+a}{\epsilon})$.
These vector operations are differentiable so $\bm{\lambda}$ can be end-to-end trained with other parts of our model.
The full derivation of this iterative algorithm is given in Appendix~\ref{appendix:derivation}.

\paragraph{Pruning Multi-vector Retrieval}
With the learned unary salience, we can naturally prune tokens for multi-vector retrieval. 
Pruning document tokens reduces the number of vectors in search index, and pruning query tokens reduces the number of searches. 
In our experiments, we control them using two pruning ratios $\beta^q$ and $\beta^d$ respectively.
For each document, we obtain the token salience using Eq.(\ref{eq:salience}) and only store the top $\beta^d$ percent of tokens in the index.
Similarly we select the top $\beta^q$ percent query tokens to perform max inner-product search.
Note that we vary these two ratios to control retrieval efficiency, and these ratios can be smaller than the sparsity ratio $\alpha^q$ and $\alpha^d$ which we use as constraints at training time.
In the refinement stage, we still use the full model with all token vectors for scoring.

\section{Experiments}

\subsection{Experimental Setup}
\label{sec:exp_setup}

\ours uses shared transformer encoder initialized from T5 version 1.1~\citep{t5}. We project token embeddings to 128 dimension and apply L2 normalization. Following  GTR~\citep{gtr}, we finetune \ours on \msmarco{} with hard negatives released by RocketQA \citep{rocketqa}. The models are trained with a batch size of 256 for 25k steps, using query sequence length of 64 and document sequence length of 256.
We train \ours with top-1 pairwise alignment.\footnote{We have trained models with other top-$k$ and top-$p$ pairwise alignments, but the MS MARCO training data favors top-1 alignment (see \Cref{sec:msmarco} for details).}

For retrieval, we pre-compute the token encodings of all the documents in the corpus, and use ScaNN~\citep{scann} to index and perform max inner-product search (MIPS). We retrieve 4,000 nearest neighbors for each query token,\footnote{Unlike ColBERT, \ours does not use pad token embeddings for retrieval. Hence, retrieving 4,000 neighbors per query token results in a similar number of retrieved candidates to ColBERT.} and return the top-1,000 after the refinement stage. We evaluate \ours on the BEIR benchmark~\citep{beir} and compare with state-of-the-art retrieval models shown in \Cref{tab:comparisons}.
Note that \ours does not rely on contrastive model pre-training~\citep{izacard2021contriever, gtr}, model-based hard negative mining~\citep{colbertv2}, or distillation~\citep{colbertv2}. We intentionally decide this simple recipe and focus on studying the impact of pairwise alignment and unary salience.

For few-shot alignment adaptation of \ours (\Cref{sec:adaptation}), we split the test data into multiple folds such that each fold contains 8 examples.
Then we find the best alignment strategy that maximizes nDCG@10 on each fold with $k\in\{1,2,4,6,8\}$ for top-$k$ and $p\in\{0.5\%,1\%,1.5\%,2\%\}$ for top-$p$.
Based on the best alignment strategy from each fold, we measure the retrieval performance on the remaining test examples with the best strategy.
We report the average ($\pm$ std.) of these test scores where the number of test scores equals the number of folds.
The average of few-shot adaptation indicates the expected performance of using few examples to choose the best alignment strategy.

\begin{table}
\small
\centering
    \renewcommand{\arraystretch}{0.8}
\resizebox{.95\linewidth}{!}{%
\begin{tabular}{l|ccccccc}
\toprule
    & \multicolumn{1}{c}{\begin{tabular}[c]{@{}c@{}}Supervision\end{tabular}} 
    & \multicolumn{1}{c}{\begin{tabular}[c]{@{}c@{}}Hard Negatives\end{tabular}}
    & \multicolumn{1}{c}{\begin{tabular}[c]{@{}c@{}}Distillation\end{tabular}}
    & \multicolumn{1}{c}{Retriever}
    & \multicolumn{1}{c}{\begin{tabular}[c]{@{}c@{}}Per-domain \end{tabular}}
    &  \multicolumn{1}{c}{\begin{tabular}[c]{@{}c@{}}\# Param.\end{tabular}} \\
\midrule

Splade$_\text{v2}$  &  \msmarco  & model-based   & \cmark    &   lexical        &       &   110M \\
ColBERT$_\text{v2}$ &  \msmarco  & model-based   & \cmark    &   multi-vector       &       &    110M  \\
GTR$_\text{xxl}$    &  Pre-train~+~\msmarco  & fixed         &       & single-vector     &           &    6B \\
\text{\textsc{Promptagator}} & Few ($\leq$ 8)  &      &       & single-vector     & \cmark    &  110M \\
\midrule
\text{\ours}$_\text{xxl}$  & \msmarco~+ Few$^*$ ($\leq$ 8) & fixed      &       & multi-vector  &       &    6B   \\

\bottomrule
\end{tabular}
}
\vspace{-0.5em}
\caption{Comparison of different retrieval models. $^*$: optionally used for alignment adaptation. 
}
\label{tab:comparisons}
\end{table}

\begin{figure}[t]
\begin{floatrow}
\capbtabbox[.65\textwidth]{%
\setlength{\tabcolsep}{2pt}
\resizebox{0.98\linewidth}{!}{%
\begin{tabular}{l|ccccc|cc}
\toprule
  & \text{BM25}       & \text{Splade}$_\text{v2}^*$    & \text{ColBERT}$_\text{v2}^*$ & \text{GTR}$_\text{xxl}$ & \text{\ours}$_\text{xxl}$ & \textsc{PTR}$^\dagger$ & \text{\ours}$_\text{xxl}^\dagger$ \\
\midrule
  MS MARCO & 18.7 & 36.8 & 39.7 & 38.8 & \textbf{40.3} & - & - \\
\hline
\midrule[0.5pt]
  \arguana      & 31.5      & 47.9      & 46.3      & \textbf{54.0}     & 33.8 & 59.4 & 47.9$^{\pm3.0}$ \\ %
  \touche       & \textbf{36.7}      & 27.2      & 26.3      & 25.6      & 34.5 & 34.5 & 50.2$^{\pm1.1}$\\ %
  \fever        & 75.3      & \textbf{78.6}      & 78.5      & 74.0      & 74.2 & 77.0 & 73.9$^{\pm4.8}$\\ %
  \climatefever & 21.3      & 23.5      & 17.6      & \textbf{26.7}      & 19.7 & 24.0 & 22.8$^{\pm2.9}$\\ %
  \scifact      & 66.5      & 69.3      & 69.3      & 66.2      & \textbf{73.1} & 65.0 & 71.4$^{\pm2.2}$\\ %
  \treccovid    & 65.6      & 71.0      & 73.8      & 50.1      & \textbf{75.8} & 75.6 & 79.3$^{\pm3.0}$\\ %
  \nfcorpus     & 32.5      & 33.4      & 33.8      & 34.2      & \textbf{35.2} & 33.4 & 33.4$^{\pm2.0}$\\ %
  \nq   & 32.9      & 52.1      & 56.2      & 56.8    & \textbf{60.5} & - & 56.6$^{\pm5.1}$\\ %
  \hotpotqa     & 60.3      & \textbf{68.4}      & 66.7      & 59.9    & 65.2 & 61.4 & 63.2$^{\pm3.3}$\\ %
  \fiqa         & 23.6      & 33.6      & 35.6      & \textbf{46.7}    & 43.5 & 46.2 & 39.9$^{\pm4.5}$\\ %
  \scidocs      & 15.8      & 15.8      & 15.4      & 16.1      & \textbf{17.1} & 18.4 & 16.3$^{\pm1.2}$\\ %
  \dbpedia      & 31.3      & 43.5      & 44.6      & 40.8   & \textbf{45.0} & 38.0 & 43.5$^{\pm2.4}$\\ %
  \quora        & 78.9      & 83.5      & 85.2      & \textbf{89.2}  & 86.0 & - & 85.3$^{\pm2.1}$\\ %
\midrule
  {\textbf{Average}} & 44.0  & 49.8      & 49.9      & 49.3      & \textbf{51.1} & - & 52.6$^{\pm3.1}$\\ %
  {\quad -- NQ / Quora}  & 41.9  & 46.6      & 46.2      & 44.9      & \textbf{47.0} & 47.8 & 49.3$^{\pm3.0}$  \\ %
\bottomrule
\end{tabular}}

}{%
  \caption{\label{tab:main}Results on MS MARCO (top; MRR@10) and the BEIR benchmark (bottom; nDCG@10). Best scores before adaptation are denoted in boldface. %
  $^*$: trained with distillation.
  \textsc{PTR}: Promptagator~\citep{promptagator}.
  $^\dagger$: uses few examples ($\leq$8) for task-specific adaptation.
  }%
}
\ffigbox[.32\textwidth]{%
  \includegraphics[width=\linewidth]{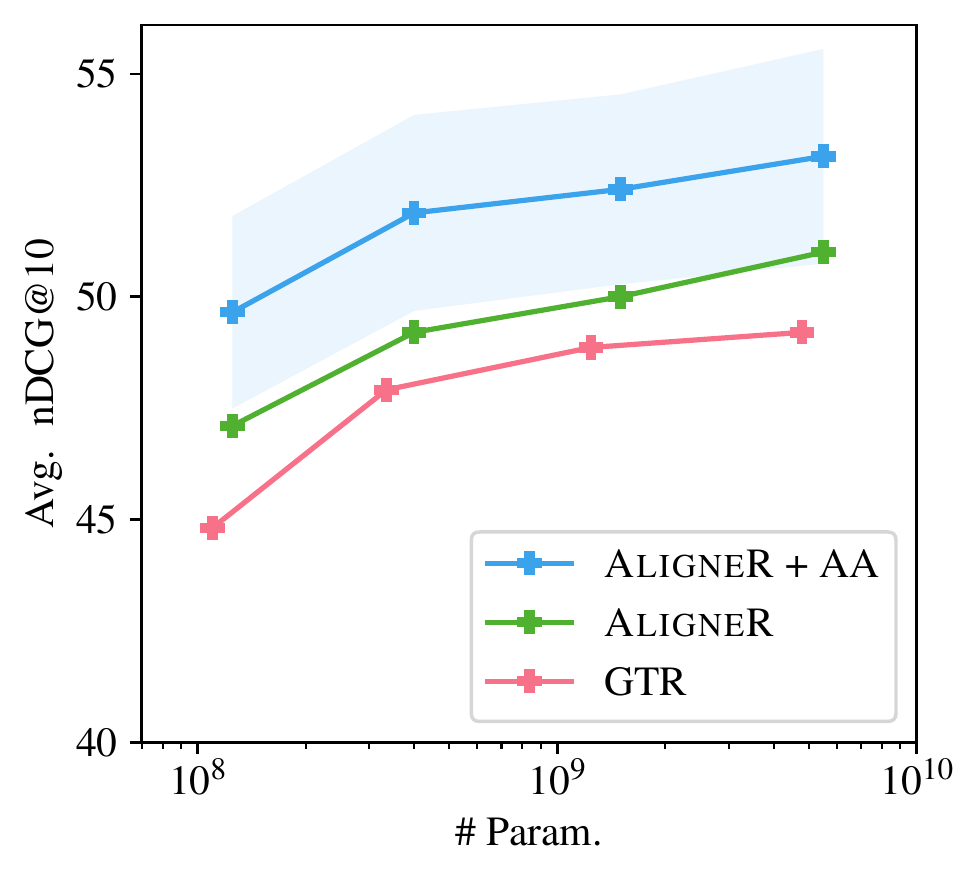}\vspace{+0.3cm}
}{%
  \caption{\label{fig:scaling}Averaged nDCG@10 on BEIR of retrieval models with different sizes.  AA: alignment adaptation.
  }%
}
\end{floatrow}
\end{figure}

\begin{figure}[t]
\begin{floatrow}
\capbtabbox[.65\textwidth]{%
    \setlength{\tabcolsep}{5pt}
  \resizebox{\linewidth}{!}{%
\begin{tabular}{lcccccccccccc}
  \toprule
  & \multicolumn{6}{c}{Top-$k$} && \multicolumn{5}{c}{Top-$p$} \\
  \cmidrule{2-7} \cmidrule{9-13}
                & 1 & 2    & 4    & 6    & 8    & 16  && $0.5\%$   & $1\%$    & $1.5\%$   & $2\%$ & $5\%$ \\  
  \midrule
                        \arguana      & 28.8 & 24.4 & 18.3 & 14.5 & 11.4 & 5.1     && 33.3 & 45.5 & \bf{48.1}   & 46.9 & 32.6  \\
                        \touche       & 34.8 & 50.0 & \bf{51.1} & 49.3 & 46.0     & 33.7      && 31.3 & 24.2 & 20.3 & 15.8 & 5.2\\
  \scifact      & 70.4 & 68.8 & 65.0 & 60.9 & 55.6 & 38.6     && \bf{71.1} & 69.4 & 67.2 & 62.7  & 39.0   \\
  \treccovid     & 68.3 & \bf{74.0} & 73.2 & 67.2 & 61.7 & 41.8 && 66.8 & 64.4 & 56.7 & 46.7 & 30.9  \\
  \fiqa         & \bf{33.4} & 30.8 & 23.8 & 19.5 & 15.5 & 8.43     && \bf{33.4} & 28.5 & 24.6 & 19.0   & 7.7  \\
  \scidocs      & 14.1 & 14.3 & 13.1 & 11.4 & 9.7 & 4.82     && 14.4 & \bf{14.9} & \bf{14.9} & 14.8   & 8.1  \\
  \dbpedia      & \bf{41.6} & 39.4 & 29.6 & 20.2 & 14.2 & 3.94     && \bf{41.6} & 41.7 & 39.9 & 36.6 & 17.0\\
\midrule
Average        & 41.6      & 43.1      & 39.2      & 36.5      & 30.6  & 19.5 && 41.7      & 41.2      & 38.8      &   34.6  & 20.1\\
\bottomrule
\end{tabular}
}

}{%
  \caption{\label{tab:alignment} {nDCG@10 on the BEIR benchmark with different $k$ and $p$ in \text{\ours}$_\text{base}$. \text{\ours}$_\text{base}$ is trained with top-$k$=1.
  } }
}
\ffigbox[.32\textwidth]{
\includegraphics[width=0.88\linewidth]{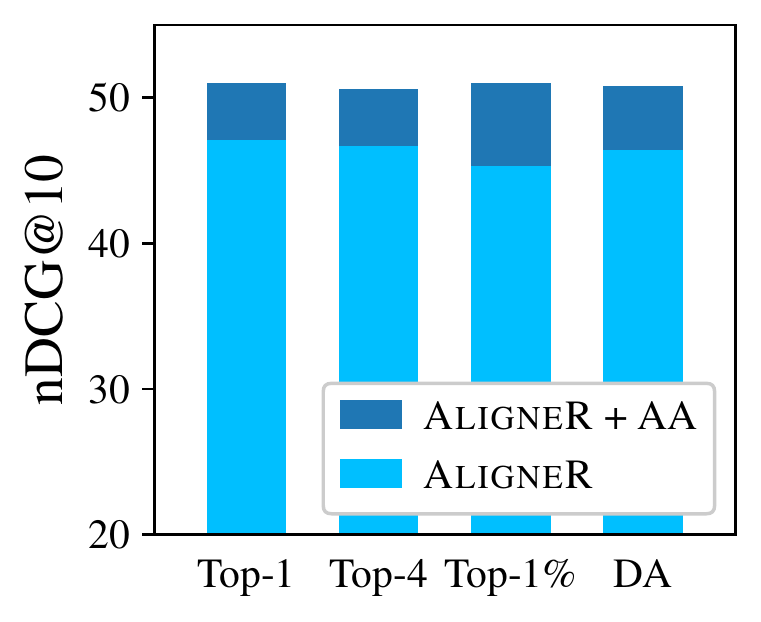}}{
    \caption{Averaged nDCG@10 on BEIR of \ours trained with different alignments. DA: \Cref{sec:diffalign}.
    }
    \label{fig:diff_align}
}
\end{floatrow}
\end{figure}

\subsection{Retrieval Accuracy}
\label{sec:doc_retrieval}
\Cref{tab:main} shows the document retrieval performance of \ours on both MS MARCO and the BEIR benchmark.
For this experiment, we do not prune any query or document tokens with unary saliences, but show their effects in \Cref{sec:efficiency} instead.
\text{\ours}$_\text{xxl}$ outperforms all baselines on MS-MARCO, showing how multi-vector retrieval models can benefit from large pretrained language models.
\text{\ours}$_\text{xxl}$ also outperforms \text{GTR}$_\text{xxl}$ on 9 out of 13 BEIR datasets and advances the retriever-only state-of-the-art (ColBERT$_\text{v2}$) by 1.2 points nDCG@10 on average.
\Cref{fig:scaling} shows that our multi-vector retriever model scales better than single-vector dual encoder GTR.

\paragraph{Alignment Adaptation}
In the rightmost column of \Cref{tab:main}, we show the effect of adapting pairwise alignment with \ours on the BEIR benchmark.
With only 8 examples for finding the proper alignment sparsity, its expected performance reaches 52.6 nDCG@10 on average.
Alignment-adapted \ours also benefits from scaling up, and consistently outperforms its non-adapted counterparts, as shown in \Cref{fig:scaling}.
The gains are further explained in \Cref{tab:alignment},  where we show individual task's performance under various alignment strategies. Although \text{\ours} is trained with top-1 alignment, top-1 is not always the best strategy at inference time.
Specifically, for \arguana{}, we observe 16 points improvement by adjusting the number of alignments proportional to the document length with $p=1.5\%$.
Other tasks such as \touche also prefer other alignment strategies, which shows that different tasks might require different sparsity.
In general, keeping the sparsity low enough is preferable and supports our hypothesis that pairwise alignments should be sparse.

We further check whether this observation holds when \ours{} is trained with other pairwise alignment strategies. \Cref{fig:diff_align} shows  \ours{} variants trained on four
alternative strategies.
We evaluate their performance with training-time alignment strategy (default) and the optimal alignment strategy selected per dataset (oracle).
While these models perform differently with their default alignments, they perform similarly after oracle alignment adaptation.

\begin{figure}[t]
    \centering
    \includegraphics[width=0.95\linewidth]{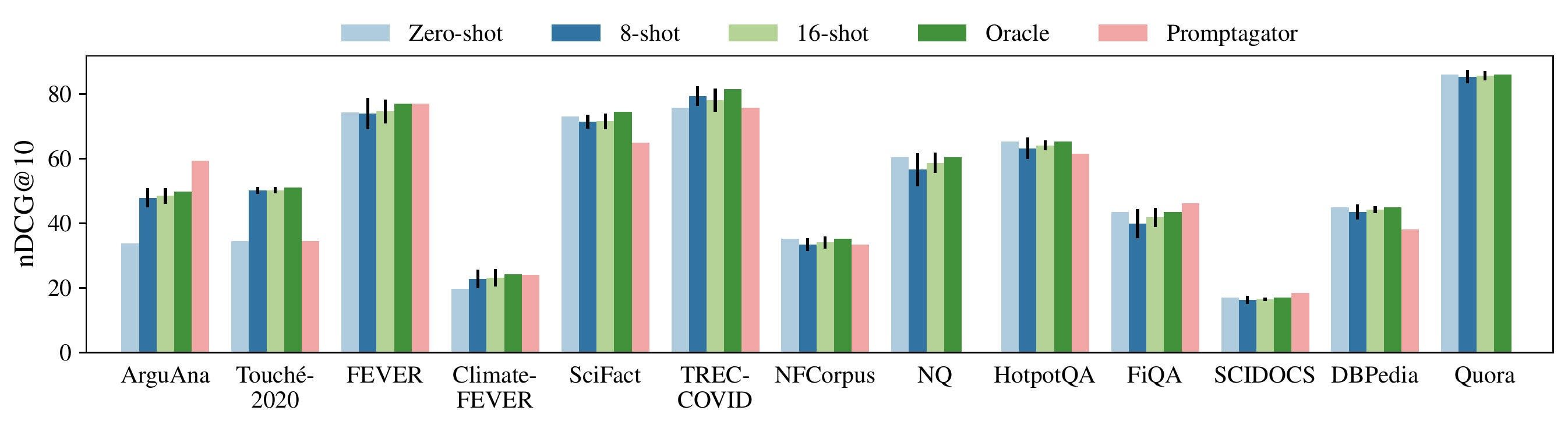}
    \vspace{-1.1em}
    \caption{\text{\ours}$_\text{xxl}$ with few-shot alignment adaptation. We report nDCG@10 on BEIR.}
    \label{fig:fewshot_beir}
\end{figure}

\Cref{fig:fewshot_beir} shows the effectiveness of few-shot alignment adaptation --- dynamically selecting task-specific alignment strategy based on a few examples.
When the default alignment (top-$k$=1) is not optimal, we can identify a good alignment strategy using only 8 examples, which significantly improves model performance on argument retrieval tasks.
Using 16 examples further improves the average score and reduces the variance.
However, when the default alignment is already optimal (top-$k$=1 is optimal for QA tasks), few-shot alignment adaptation hurts performance due to the variance of our few-shot method.
Nevertheless, \ours outperforms Promptagator~\citep{promptagator}, another few-shot retrieval baseline, in 6 out of 11 datasets.

\begin{figure}[t]
\vspace{-0.7em}
\begin{floatrow}
\ffigbox[.33\textwidth]{
    \includegraphics[width=\linewidth]{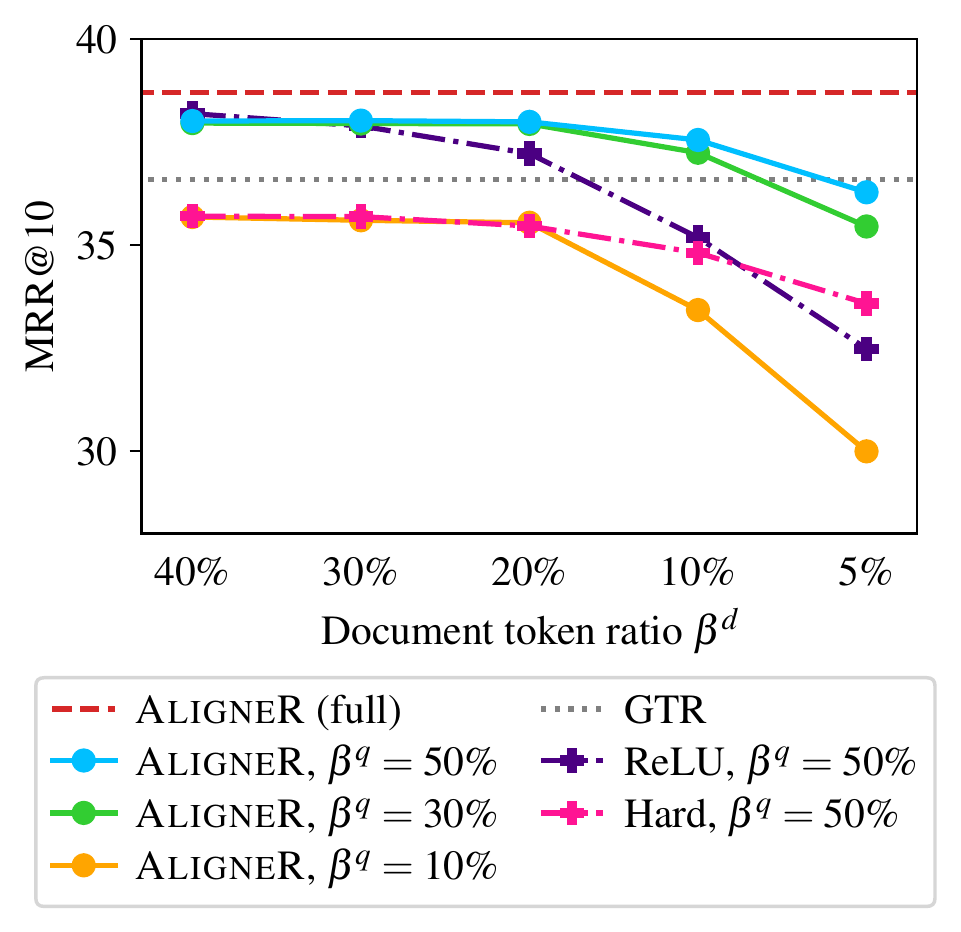} \vspace{-1.8em}
}{
    \caption{\ours with unary salience on MS~MARCO. $\beta^q$ and $\beta^d$ are ratios to prune query and document tokens, respectively.}
    \label{fig:salience_msmarco}
}
\ffigbox[.65\textwidth]{
    \includegraphics[width=0.99\linewidth]{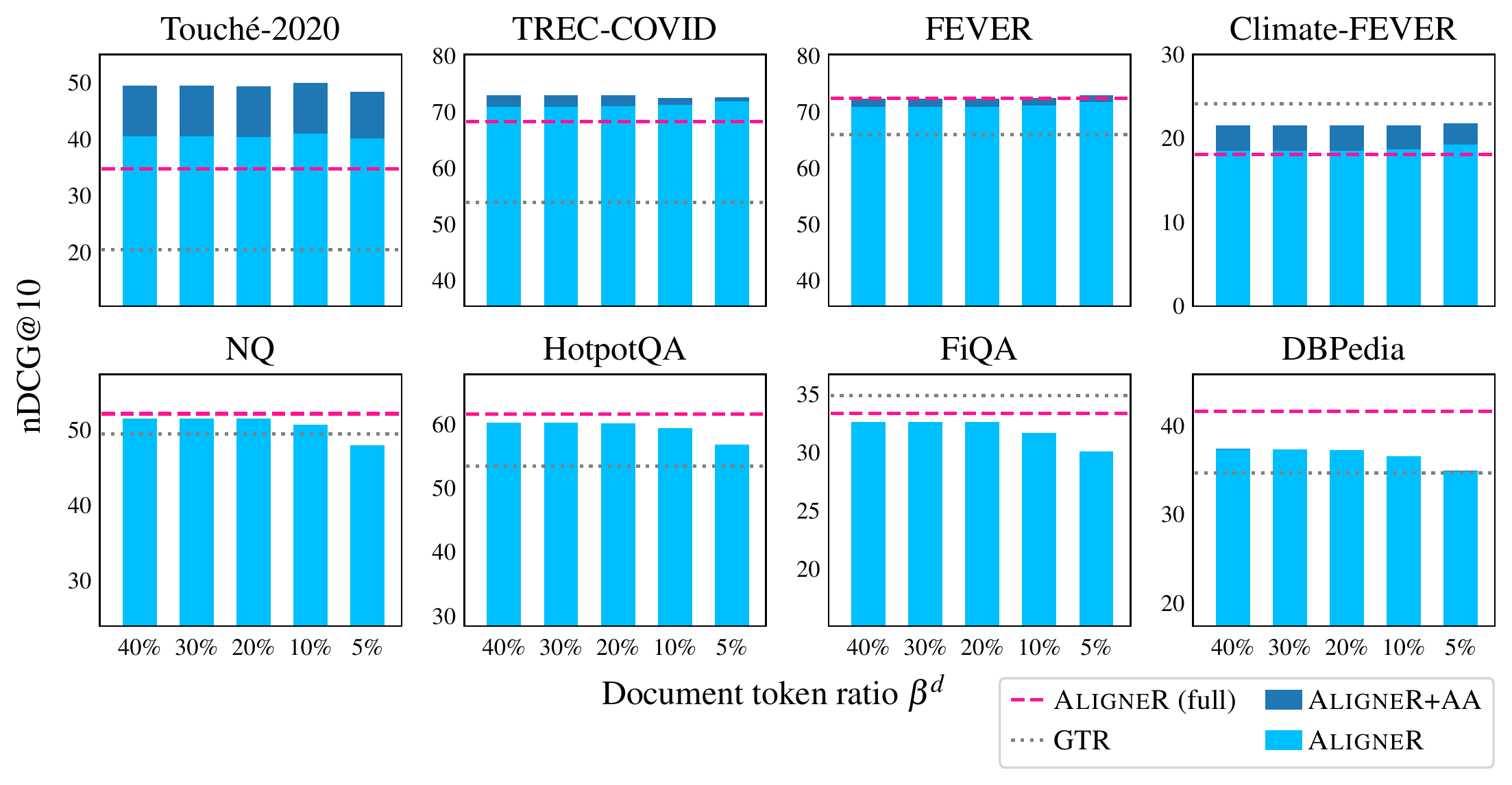} \vspace{-1.em}
}{
    \caption{\ours with unary salience on BEIR. We set query pruning ratio $\beta^q=50\%$ and vary document pruning ratio $\beta^d$. We omit datasets with small corpora. We also report the performance of \ours with alignment adaptation (AA).} %
    \label{fig:salience_beir}
}
\end{floatrow} %
\end{figure}

\subsection{Retrieval Efficiency}
\label{sec:efficiency}

The next experiment shows how \ours{}'s unary salience impacts retrieval efficiency. We train \text{\ours}$_\text{base}$ with salience sparsity ratios $\alpha^q=50\%$ and $\alpha^d=40\%$ based on empirical performance. The gating variables are optimized with $\epsilon=0.002$. At retrieval time, we prune query and document tokens with ratios $\beta^q$ and $\beta^d$ (\Cref{sec:coarse}).

\Cref{fig:salience_msmarco} shows the \ours{} performance on \msmarco{} with various pruning ratios. When pruned at the same ratio as training ($\beta^q=50\%$ and $\beta_d=40\%$), the model is close to a full \ours model without pruning (MRR@10 38.1 vs. 38.8), but greatly saves the computation cost. We can further prune tokens by adjusting $\beta^d$ and $\beta_q$. The model achieves 37.3 MRR@10 with is $\beta^d=10\%$, i.e. it remains accurate with only 10\% of the original index size. Decreasing the query pruning ratio $\beta^q$ to 30\% does not sacrifice performance too much, although deceasing $\beta^q$ to $10\%$ leads to worse performance. \Cref{fig:salience_msmarco} also compares \ours{}'s entropy-regularized linear program (Eq.~\ref{eq:opt}) with alternative methods. With just a ReLU gate and no sparsity constraints (`ReLU' in \Cref{fig:salience_msmarco}), the model performance retains good  when $\beta^d=40\%$, but drops significantly for smaller $\beta^d$.  
Removing the entropy regularization in Eq.~\ref{eq:opt} leads to simply selecting the hard top-$k$ tokens with the highest predicted salience (`Hard' in \Cref{fig:salience_msmarco}). The hard top-$k$ solution has worse performance for all $\beta^d$.

\ours{}'s salience estimation also generalizes to other retrieval datasets. As shown in \Cref{fig:salience_beir}, pruning with $\beta_d=10\%$ with  $\beta^q=50\%$ causes minimal performance decrease for a majority of BEIR datasets. We even observe performance increase for \touche, as the model can only retrieve salient tokens after pruning. Besides, we show that alignment adaptation can be combined with pruning, resulting in an effective yet efficient retrieval model.

\subsection{Interpretability}\label{sec:interpretability}

\begin{table}[t]
    \centering
    \footnotesize
    \renewcommand{\arraystretch}{0.7}
\resizebox{\linewidth}{!}{%
    \begin{tabular}{lm{5.4in}}
    \toprule

        Query& \hlr{what}{0} \hlr{happens}{10} \hlr{when}{0} \textcolor{red}{\textbf{\hlr{stop}{46}}} \textcolor{ForestGreen}{\textbf{\hlr{drinking}{28}}} \textcolor{blue}{\textbf{\hlr{alcohol}{46}}} \\\midrule
        Doc.& \textcolor{blue}{\textbf{\hlr{alcohol}{81}}} \hlr{.}{0} \hlr{symptoms}{65} \hlr{of}{0} \hlr{alcohol}{79} \textcolor{red}{\textbf{\hlr{withdrawal}{76}}} \hlr{may}{0} \hlr{begin}{0} \hlr{from}{0} \hlr{4}{0} \hlr{to}{0} \hlr{12}{0} \hlr{hours}{67} \hlr{after}{0} \hlr{you}{0} \hlr{cut}{74} \hlr{down}{0} \hlr{or}{0} \hlr{stop}{73} \textcolor{ForestGreen}{\textbf{\hlr{drinking}{77}}} \hlr{,}{0} \hlr{or}{0} \hlr{as}{0} \hlr{long}{0} \hlr{as}{0} \hlr{several}{0} \hlr{days}{0} \hlr{after}{0} \hlr{the}{0} \hlr{last}{0} \hlr{drink}{73} \hlr{,}{0} \hlr{and}{0} \hlr{can}{0} \hlr{last}{0} \hlr{}{0} \hlr{a}{0} \hlr{few}{0} \hlr{days}{0} \hlr{.}{0} \hlr{they}{0} \hlr{can}{0} \hlr{range}{0} \hlr{from}{0} \hlr{mild}{0} \hlr{to}{0} \hlr{life}{0} \hlr{-}{0} \hlr{threatening}{69} \hlr{.}{0} \hlr{1}{0} \hlr{mild}{0} \hlr{withdrawal}{84} \hlr{symptoms}{63} \hlr{may}{0} \hlr{include}{0} \hlr{:}{0} \hlr{2}{0} \hlr{intense}{0} \hlr{worry}{70} \hlr{.}{0} \hlr{3}{0} \hlr{nausea}{64} \hlr{or}{0} \hlr{vomiting}{0} \hlr{.}{0} \hlr{4}{0} \hlr{}{0} \hlr{s}{0} \hlr{hak}{70} \hlr{iness}{0} \hlr{.}{0} \hlr{5}{0} \hlr{sweat}{76} \hlr{ing}{0} \hlr{.}{0}  \\\toprule
        
        Query & \hlr{where}{16} \hlr{is}{0} \hlr{the}{0} \textcolor{red}{\textbf{\hlr{heart}{46}}} \hlr{in}{0} \hlr{the}{0} \textcolor{ForestGreen}{\textbf{\hlr{human}{21}}} \textcolor{blue}{\textbf{\hlr{body}{22}}} \\\midrule
        Doc. & \textcolor{red}{\textbf{\hlr{heart}{78}}} \hlr{the}{0} \hlr{heart}{79} \hlr{is}{0} \hlr{}{0} \hlr{a}{0} \textcolor{ForestGreen}{\textbf{\hlr{muscular}{61}}} \textcolor{blue}{\textbf{\hlr{organ}{57}}} \hlr{in}{0} \hlr{most}{0} \hlr{animals}{0} \hlr{,}{0} \hlr{which}{0} \hlr{pumps}{59} \hlr{blood}{57} \hlr{through}{0} \hlr{the}{0} \hlr{blood}{0} \hlr{vessels}{57} \hlr{of}{0} \hlr{the}{0} \hlr{circul}{58} \hlr{atory}{0} \hlr{system}{0} \hlr{.}{0} \hlr{[}{0} \hlr{1}{0} \hlr{]}{0} \hlr{blood}{56} \hlr{provides}{60} \hlr{the}{0} \hlr{body}{0} \hlr{with}{0} \hlr{oxygen}{53} \hlr{and}{0} \hlr{nutrients}{0} \hlr{,}{0} \hlr{as}{0} \hlr{well}{0} \hlr{as}{0} \hlr{assists}{0} \hlr{in}{0} \hlr{the}{0} \hlr{removal}{0} \hlr{of}{0} \hlr{metabolic}{0} \hlr{waste}{0} \hlr{s}{0} \hlr{.}{0} \hlr{[}{0} \hlr{2}{0} \hlr{]}{0} \hlr{in}{0} \hlr{humans}{51} \hlr{,}{0} \hlr{the}{0} \hlr{heart}{71} \hlr{is}{0} \hlr{located}{0} \hlr{between}{0} \hlr{the}{0} \hlr{}{0} \hlr{lungs}{57} \hlr{,}{0} \hlr{in}{0} \hlr{the}{0} \hlr{middle}{0} \hlr{compartment}{0} \hlr{of}{0} \hlr{the}{0} \hlr{chest}{55} \hlr{.}{0} \hlr{[}{0} \hlr{3}{0} \hlr{]}{0} \\
        
         \bottomrule
    \end{tabular}
}
    \caption{Examples of the pairwise alignment and unary salience  learned by \ours. Three most salient query tokens and their top-1 pairwise-aligned document tokens are indicated with the same color. We highlight top 50\% query tokens and 20\% document tokens according to their salience.  
    }
    \label{tab:salience_example}
\end{table}
\Cref{tab:salience_example} shows examples of the pairwise alignment and unary salience learned by \ours. The model aligns query tokens to contexually similar tokens, but not necessarily identical tokens. The salience features are also highlighted in \Cref{tab:salience_example}. Important noun phrases and verbs are usually assigned higher salience, which is consistent with human intuition. We show more examples of alignments for different tasks in the \Cref{app:qualitative}. In general, we observe question answering tasks usually require fewer alignments for each query token, while other tasks that require a broad understanding of the document favor larger number of alignments.

\section{Related Work}
\label{sec:related_work}
Recent research on information retrieval often improves the retrieval accuracy with contrastive pretraining~\citep{gtr, izacard2021contriever, oguz-etal-2022-domain}, model-based hard negative mining~\citep{xiong2020approximate, lu-etal-2021-multi, rocketqa} and knowledge distillation~\citep{colbertv2, zhang2022adversarial, DBLP:conf/aistats/ReddiPMRYKVK21}.
Retrieval efficiency is improved via quantization~\citep{colbertv2} or lower-dimensional vectors~\citep{colberter}.
These improvements are orthogonal to this work.

Term importance and salience have a long history in information retrieval: from
term frequency (\textit{tf}) and inverse document frequency (\textit{idf}), to recent BERT-based importance measures such as DeepCT~\citep{dai2020deepct_short}, SPARTA~\citep{zhao2021sparta} and Splade~\citep{formal21splade, spladev2}. These works mostly focus on sparse lexical retrieval and learn term weights for sparse bag-of-words representations. Term importance in multi-vector dense retrieval is less explored. Our work is probably most related to a recent work from~\cite{colberter}, which prunes ColBERT by predicting salience scores from a word's embedding with a ReLU gate and \text{L1-norm} regularization.

Recently, Promptagator~\citep{promptagator} points out the importance of using a few annotated examples to adapt to a new retrieval task. Promptagator achieves few-shot task adaptation via query generation ~\citep{MaKYHM21,ex2, DaiCZARGG22} using large language models~\citep{tzero, gpt3, flan}, which has high inference cost. \ours is more versatile and can be fast adapted to a new task via few-shot alignment adaptation.

\section{Conclusion}
In this paper, we introduce \ours, a novel sparse alignment method for multi-vector document retrieval.
We first formulate different retrieval models with token-level sparse alignments and propose \ours to tackle the limitations of existing models.
Specifically, \ours uses pairwise alignments and unary saliences that allow us to adapt to different tasks and prune unimportant tokens, respectively.
As a result, we achieve strong performance on both zero-shot and few-shot document retrieval tasks while drastically improving the run-time and storage complexity of multi-vector retrieval.
With its interpretable alignments and better performance with large language models, we envision that our multi-vector retrieval model can serve as a strong standalone retriever in the future. %

\section{Author Contributions}
{\em Yujie Qian} and {\em Vincent Y. Zhao} are the leading authors of this work who designed the model and experiments. All the authors have contributed to the code, experiments, and paper writing. {\em Sai Meher Karthik Duddu} conducted the interpretability analysis. {\em Jinhyuk Lee}, {\em Zhuyun Dai}, {\em Iftekhar Naim}, and {\em Tao Lei} advised on the research direction. {\em Tao Lei} and {\em Siddhartha Brahma} proposed the algorithm for entropy-regularized linear programming. {\em Vincent} initiated the project and proposed the modeling framework. {\em Vincent} and {\em Tao} were the co-hosts of {\em Yujie}'s internship.

\newpage
\bibliography{references}
\bibliographystyle{iclr2023_conference}

\newpage
\appendix
\section{Appendix}

\subsection{Derivation of the Iterative Updates}
\label{appendix:derivation}

We present the derivation of Eq.\ref{eq:iterations} for solving optimization problem~(\ref{eq:opt}) in Section~\ref{sec:coarse}.
The maximization problem~(\ref{eq:opt}) can be written as an equivalent minimization problem:
\begin{align}
 & \max_{\bm{\lambda}} \ \ \vs^\top \bm{\lambda} + \epsilon H(\bm{\lambda}) \nonumber \\
\Longleftrightarrow\qquad & \min_{\bm{\lambda}} \ \ -\vs^\top \bm{\lambda} - \epsilon H(\bm{\lambda}) \nonumber \\
\Longleftrightarrow\qquad & \min_{\bm{\lambda}} \ \ -\vs^\top \bm{\lambda} - \epsilon H(\bm{\lambda}) - \epsilon \bm{1}^\top \bm{\lambda} \label{eq:minobj}\\
\text{s.t.}\  & \bm{1}^\top\bm{\lambda} = k, \ \  \lambda_i\in [0, 1],\ \ i=1,\dots,m. \nonumber
\end{align}
Note the term $\epsilon \bm{1}^\top \bm{\lambda}$ will be a constant $\epsilon\times k$, but we include it in the minimization object to make our derivation simpler later.

Now, let $a\in\R$ and $\vb\in\R^m$ be the Lagrangian variables corresponding to the linear constraints $\bm{1}^\top\bm{\lambda} = k$ and $\lambda_i \leq 1\ \forall i\ $.\footnote{$\lambda_i \geq 0\ \forall i\ $ is already implied by the entropy term $H(\bm{\lambda})$ in the objective.}
The minimization problem is equivalent to its Lagrangian expression:
\begin{align}
\min_{\lambda\in\R^m}\ \max_{a\in\R, \vb\leq \bm{0}} \ -\vs^\top \bm{\lambda} - \epsilon H(\bm{\lambda}) - \epsilon \bm{1}^\top \bm{\lambda} + a (k-\bm{1}^\top\bm{\lambda}) + \bm{b}^\top (\bm{1} - \bm{\lambda})
\label{eq:minmax}
\end{align}
The objective function~(\ref{eq:minobj}) is strongly convex and the solution space of $\bm{\lambda}$ is a convex set. As a result, strong duality holds and we can instead solve the dual problem that exchanges the $\min$ and $\max$ operators in (\ref{eq:minmax})
\begin{align}
\max_{a\in\R, \vb\leq \bm{0}}\ \min_{\lambda\in\R^m}\ -\vs^\top \bm{\lambda} - \epsilon H(\bm{\lambda}) - \epsilon \bm{1}^\top \bm{\lambda} + a (k-\bm{1}^\top\bm{\lambda}) + \bm{b}^\top (\bm{1} - \bm{\lambda})
\label{eq:maxmin}
\end{align}
The optimal solution $(a, \vb, \bm{\lambda})$ must have the Karush-Kuhn-Tucker (KKT) conditions hold~\citep{kuhn2014nonlinear}, namely
\begin{align*}
& \frac{\partial\left(-\vs^\top \bm{\lambda} - \epsilon H(\bm{\lambda}) + \epsilon \bm{1}^\top \bm{\lambda} + a (k-\bm{1}^\top\bm{\lambda}) + \bm{b}^\top (\bm{1} - \bm{\lambda})\right)}{\partial \bm{\lambda}} = 0 \\
\Longleftrightarrow\qquad & \bm{\lambda} = \exp\left(\frac{\vs + a +\vb}{\epsilon}\right)
\quad \Longleftrightarrow\quad \lambda_i = \exp\left(\frac{s_i + a+\vb_i}{\epsilon}\right) \ \ \forall i=1,\dots,m
\end{align*}
Substituting $\bm{\lambda}$ using the above equation in (\ref{eq:maxmin}), the dual problem now has a simple form:
\begin{align*}
\max_{a\in\R, \vb\leq \bm{0}} k\cdot a + \bm{1}^\top\vb - \bm{1}^\top \exp\left( \frac{\vs + a +\vb}{\epsilon}\right)
\end{align*}
We can solve this problem using coordinate descent~\citep{wright2015coordinate} by successively maximizing the function with either $a$ or $\vb$ fixed.
This leads to the iterative updates (Eq.\ref{eq:iterations}) described in Section~\ref{sec:coarse}.
\begin{equation}
\begin{split}
    a' &= \epsilon \ln(k) - \epsilon \ln\left\lbrace\sum_i \exp\left(\frac{s_i + b_i}{\epsilon}\right)\right\rbrace\\
    b'_i &= \min(-s_i - a', 0) \nonumber
\end{split}
\end{equation}

\paragraph{Discussion}
In short, we solve the dual problem of optimization~(\ref{eq:opt}) by performing coordinate decent of the dual variables $a$ and $\vb$.
That is, we find the optimal $a$ that maximizes the dual objective given a fixed $\vb$, and vice versa.

This iterative algorithm is also closely related to the Sinkhorn algorithm of Optimal Transport (OT). 
In fact, Sinkhorn algorithm solves the entropy-regularized version of Optimal Transport~\citep{cuturi2013sinkhorn}.
However, our work concerns an different optimization instance. While OT solves a transportation problem where the solution space is defined with the marginal constraints over the rows and columns of a transportation matrix, our optimization problem is constrained with a total budget ($\sum_i \lambda_i = k$) and upper bounds ($\lambda_i \leq 1\ \forall i$). 
This leads to different iterative updates.

\subsection{Differentiable Alignment with Sparsity Constraints}
\label{sec:diffalign}
Besides the Top-$k$ and Top-$p$ alignments in \Cref{sec:adaptation}, we also explore a differentiable pairwise alignment with sparsity contraints (DA). Both Top-$k$ adn Top-$p$ are doing hard selection of alignments, i.e., $\tilde{\mA}_{i,j}$ is either 1 or 0. We relax it by introducing soft sparsity constraints. Similar to our formulation for unary salience (\Cref{sec:coarse}), we determine the alignment $\tilde{\mA}$ by the following optimization problem:
\begin{equation}
    \begin{split}
    \max_{\mA} & ~ \langle \mS, \mA \rangle + \epsilon H(\mA) \\
    \text{s.t. } & \sum_{j} \mA_{i,j} = k, \ i=1,\dots,n \\
    & \mA_{i,j} \in [0, 1],\ \ i=1,\dots,n,\  j=1,\dots,m
\end{split}
\label{eq:diffalign}
\end{equation}
where  $H(\cdot)$ is the elementwise entropy function and $\epsilon > 0$ is a small constant. We constrain the sum of each row of $\tilde{\mA}$ to equal $k$. When $\epsilon=0$, the solution of Eq.~\ref{eq:diffalign} is the same as Top-$k$.  When $\epsilon>0$, the entropy term makes the optimization problem  strongly concave, which can be solved by the same algorithm in \Cref{appendix:derivation}. The solution is differentiable, thus can be trained end-to-end in our model.

\subsection{Qualitative Analysis}
\label{app:qualitative}

\begin{table}[h]
\centering
\footnotesize
\resizebox{\linewidth}{!}{%
\begin{tabular}{p{0.6in}p{1.1in}p{3.4in}}
\toprule
\textbf{Dataset} & \textbf{Query} & \textbf{Gold Document} \\
\midrule
\quora & what is the \textcolor{blue}{best} birthday gift for a friend? & \textcolor{red}{what\textsuperscript{(3)}} \textcolor{red}{is\textsuperscript{(2)}} \textcolor{red}{a\textsuperscript{(4)}} \textcolor{red}{good\textsuperscript{(1)}} birthday gift for a friend? \\
\midrule
MS~MARCO (dev) & when would you use a \textcolor{blue}{fathom} measurement & a \textcolor{red}{fathom\textsuperscript{(1)}} is a unit of length in the imperial and the u.s. customary systems equal to 6 \textcolor{red}{feet\textsuperscript{(3)}} (1.8288 \textcolor{red}{m\textsuperscript{(4)}}), used especially for measuring the depth of water. there are two yards (6 feet) in an imperial \textcolor{red}{fathom\textsuperscript{(2)}}. \\
\midrule
\touche & should animals be used for \textcolor{blue}{scientific} or commercial testing? & animal testing should not be allowed\ldots \textit{[truncated]}\ldots  albeit the non-precocious mistakes of \textcolor{red}{scientists\textsuperscript{(2)}}. also\ldots \textit{[truncated]}\ldots  skeptic of the \textcolor{red}{scientist\textsuperscript{(3)}} in question's abilities \ldots \textit{[truncated]}\ldots continuous use if animals for \textcolor{red}{clinical\textsuperscript{(4)}} and basic research." \ldots \textit{[truncated]}\ldots majority of the \textcolor{red}{scientific\textsuperscript{(1)}} community thinks on this issue, \ldots \\  
\bottomrule
\end{tabular}
}
\caption{\label{tab:align_example} {Examples of pairwise alignment with the top-$k$ value up to 4 for the Quora, MS MARCO, and \touche datasets. Query tokens being aligned are shown in blue, and corresponding aligned document tokens are shown in red. The superscript on the document token $(k)$ indicates top-$k$ alignment. We notice that the top-1 alignment quality is generally good across all three tasks. However, larger value of $k$ results in spurious irrelevant alignments for Quora and MS MARCO, while remains fairly useful for \touche.}}
\end{table} 
\Cref{tab:align_example} shows examples of top-$k$ pairwise alignments of a query token (highlighted in blue) to the corresponding document tokens for several different tasks. 
For question-answering (e.g., MS MARCO) and duplicate question retrieval (Quora), fewer alignments seem to be preferable, and as $k$ increases, we start to see spurious alignments to unrelated documents tokens. For argument retrieval tasks such as \touche, on the other hand, larger value of $k$ tends to provide useful semantically relevant alignments (e.g., \textit{scientific} vs \textit{clinical}). These qualitative examples provide intuitive insights regarding why different alignment strategies are helpful for different tasks, and why alignment adaptation is necessary. 

\subsection{Results on \msmarco}
\label{sec:msmarco}
\begin{table}[ht]
    \centering
    \small
    \begin{tabular}{lccccc}
    \toprule
        Model                       & MRR@10 & Recall@1000 \\\midrule %
        BM25                        &  18.7    &  85.7   \\ %
        SPLADE$_{\text{v2}}$        & 36.8    &  97.9   \\ %
        \midrule 
        DPR                         & 31.1    &  95.2   \\ %
        GTR$_\text{base}$           & 36.6    &  98.3   \\ %
        GTR$_\text{large}$          & 37.9    &  \textbf{99.1}   \\ %
        GTR$_\text{xl}$             & 38.5    &  98.9   \\ %
        GTR$_\text{xxl}$            & 38.8    &  99.0   \\ %
        \midrule
        ColBERT                     & 36.0    &  96.8   \\ %
        ColBERT$_{\text{v2}}$       & 39.7    &  98.4   \\ %
        COIL                        & 35.5    &  96.3   \\ %
        ME-BERT                     & 33.4    &  --     \\ %
        \midrule
        \midrule
        \oursbase                   & 38.8   & 97.8   \\ %
        \ourslarge                  & 39.4   & 98.3   \\ %
        \oursxl                     & 39.9   & 98.4   \\ %
        \oursxxl                    & \textbf{40.3}   & 98.7   \\ %
        \midrule
        {\textsc{AligneR}$_\text{base}$} (top-$4$)      & 37.1   & 97.5   \\ %
        {\textsc{AligneR}$_\text{base}$} (top-$1\%$)    & 38.8   & 97.6   \\ %
        {\textsc{AligneR}$_\text{base}$} (DA)    & 37.8   & 97.4   \\
        
    \bottomrule
    \end{tabular}
    \caption{Retrieval performance on \msmarco. The top half shows baselines from previous work. The botton half shows different \ours models. DA: differential alignment. See \Cref{sec:diffalign}}
    \label{tab:msmarco}
\end{table}
\Cref{tab:msmarco} shows the retrieval performance of \ours and previous models on the MS MARCO dev set. We deliberately kept the training configuration of \ours relatively simpler (e.g., no distillation or model-based hard negatives).
However, \ours still achieves the best MRR@10 simply because of scaling to larger pretrained language models. We have also trained \ours with other alignment strategies, such as top-4, top-1\%, and DA (\Cref{sec:diffalign}). However, the results suggest top-1 is favorable in MS MARCO.

\subsection{Full Result Tables on BEIR}
\Cref{full_bier_base,full_beir_large,full_beir_xl,full_beir_xxl} presents complete results of \ours's performance on the BEIR datasets initialized from T$5$ base, large, XL, and XXL checkpoints. 
We set $k=1$ during training, and show results across different inference-time alignment strategies (both top-$k$ and top-$p$).
As expected, model accuracy improves as we scale to larger models. 
Moreover, we observe similar benefits of alignment adaptation across all the different model sizes.

\begin{table*}[h]
  \small
  \centering
\begin{tabular}{clcccccccccc}
  \toprule
  \multicolumn{2}{c}{\multirow{2}{*}{\text{\ours}$_\text{base}$}}& \multicolumn{5}{c}{Top-$k$} & & \multicolumn{4}{c}{Top-$p$}\\ %
  \cmidrule{3-7}\cmidrule{9-12}
  &              & $1^*$& $2$    & $4$    & $6$    & $8$    && $0.5\%$   & $1\%$    & $1.5\%$   & $2\%$  \\  
  \midrule
  \multirow{2}{*}{A.R.} &\arguana       & 28.8 & 24.4 & 18.3 & 14.5 & 11.4 && 33.3 & 45.5 & 48.1 & 46.9 \\ %
  &\touche                              & 34.8 & 50.0 & 51.1 & 49.3 & 46.0 && 31.3 & 24.2 & 20.3 & 15.8 \\ %
  \cmidrule(lr){1-2}
  \cmidrule(lr){1-2}
  \multirow{3}{*}{F.C.} &\fever         & 72.4 & 75.0 & 68.3 & 57.6 & 49.0 && 72.5 & 55.0 & 44.9 & 29.9 \\ %
  &\climatefever                        & 18.1 & 20.8 & 23.0 & 23.0 & 22.7 && 18.2 & 13.7 & 13.8 & 10.8 \\ %
  &\scifact                             & 70.4 & 68.8 & 65.0 & 60.9 & 55.6 && 71.1 & 69.4 & 67.2 & 62.7 \\ %
  \cmidrule(lr){1-2}
  \cmidrule(lr){1-2}
  \multirow{5}{*}{Q.A.} &\nq            & 52.2 & 48.3 & 36.3 & 26.6 & 19.9 && 52.2 & 49.2 & 43.5 & 36.3 \\ %
  &\hotpotqa                            & 61.7 & 58.6 & 36.0 & 21.9 & 13.9 && 61.7 & 60.1 & 54.3 & 47.3 \\ %
  &\fiqa                                & 33.4 & 30.8 & 23.8 & 19.5 & 15.5 && 33.4 & 28.5 & 24.6 & 19.0 \\ %
  &\bioasq                              & 49.6 & 45.8 & 37.4 & 30.8 & 24.5 && 47.4 & 37.7 & 31.9 & 25.0 \\ %
  &\nfcorpus                            & 34.0 & 33.2 & 32.0 & 29.9 & 27.8 && 33.6 & 31.7 & 30.5 & 28.6 \\ %
  \cmidrule(lr){1-2}
  \multirow{4}{*}{MISC.} &\treccovid    & 68.3 & 74.0 & 73.2 & 67.2 & 61.7 && 66.8 & 64.4 & 56.7 & 46.7 \\ %
  &\scidocs                             & 14.1 & 14.3 & 13.1 & 11.4 & 9.7  && 14.4 & 14.9 & 14.9 & 14.8 \\ %
  &\dbpedia                             & 41.6 & 39.4 & 29.6 & 20.2 & 14.2 && 41.6 & 41.7 & 39.9 & 36.6 \\ %
  &\quora                               & 82.3 & 64.9 & 30.6 & 13.3 & 6.3  && 82.3 & 82.3 & 82.3 & 82.3 \\ %
  \midrule
  & Average                                 & 47.3 & 46.3 & 38.4 & 31.9 & 27.0 && 47.1 & 44.2 & 40.9 & 35.9  \\ %
  
\bottomrule
\end{tabular}
\caption{\label{full_bier_base}nDCG@10 on the BEIR benchmark with different $k$ and $p$ in \text{\ours}$_\text{base}$. $^*$: alignment strategy during training ($k=1$).}
\end{table*}

\begin{table*}[ht]
  \small
  \centering
\begin{tabular}{clcccccccccc}
  \toprule
  \multicolumn{2}{c}{\multirow{2}{*}{\text{\ours}$_\text{large}$}}& \multicolumn{5}{c}{Top-$k$} & & \multicolumn{4}{c}{Top-$p$}\\ %
  \cmidrule{3-7}\cmidrule{9-12}
  &              & $1^*$& $2$    & $4$    & $6$    & $8$    && $0.5\%$   & $1\%$    & $1.5\%$   & $2\%$  \\
  \midrule
  \multirow{2}{*}{A.R.} &\arguana       & 29.5 & 25.2 & 19.5 & 16.0 & 13.6 && 33.2 & 44.5 & 47.9 & 46.8 \\ %
  &\touche                              & 36.7 & 47.5 & 53.0 & 53.4 & 52.0 && 32.5 & 25.4 & 20.7 & 16.3 \\ %
  \cmidrule(lr){1-2}
  \cmidrule(lr){1-2}
  \multirow{3}{*}{F.C.} &\fever         & 72.9 & 75.2 & 69.6 & 61.8 & 53.6 && 72.8 & 52.8 & 43.0 & 29.3 \\ %
  &\climatefever                        & 18.6 & 20.7 & 23.3 & 23.4 & 23.5 && 18.6 & 13.1 & 14.2 & 11.4 \\ %
  &\scifact                             & 71.5 & 70.7 & 69.5 & 67.7 & 64.1 && 72.2 & 70.6 & 69.6 & 66.8 \\ %
  \cmidrule(lr){1-2}
  \cmidrule(lr){1-2}
  \multirow{5}{*}{Q.A.} &\nq            & 57.2 & 52.8 & 43.4 & 36.3 & 31.2 && 57.2 & 53.7 & 47.5 & 41.0 \\ %
  &\hotpotqa                            & 63.6 & 62.6 & 44.8 & 32.1 & 24.4 && 63.6 & 61.8 & 55.6 & 48.5 \\ %
  &\fiqa                                & 39.4 & 35.6 & 30.4 & 26.2 & 23.4 && 39.4 & 33.3 & 28.4 & 22.8 \\ %
  &\bioasq                              & 53.3 & 50.7 & 43.1 & 36.8 & 31.6 && 49.2 & 38.2 & 33.2 & 27.4 \\ %
  &\nfcorpus                            & 35.5 & 33.9 & 32.5 & 31.2 & 29.6 && 34.6 & 32.1 & 30.8 & 29.6 \\ %
  \cmidrule(lr){1-2}
  \multirow{4}{*}{MISC.} &\treccovid     & 71.9 & 79.4 & 77.3 & 74.4 & 69.3 && 70.0 & 66.1 & 59.7 & 52.4 \\ %
  &\scidocs       & 15.3 & 15.5 & 14.9 & 13.6 & 12.5 && 15.4 & 15.8 & 16.0 & 15.8 \\ %
  &\dbpedia      & 43.5 & 41.9 & 34.7 & 29.0 & 24.3 && 43.5 & 43.5 & 41.5 & 37.4 \\ %
  &\quora                               & 84.5 & 75.5 & 46.4 & 20.9 & 8.4  && 84.5 & 84.5 & 84.5 & 84.5 \\ %
  \midrule
  & Average                                & 49.5 & 49.1 & 43.0 & 37.3 & 33.0 && 49.0 & 45.4 & 42.3	& 37.9 \\ %
  
\bottomrule
\end{tabular}
\caption{\label{full_beir_large}nDCG@10 on the BEIR benchmark with different $k$ and $p$ in \text{\ours}$_\text{large}$. $^*$: alignment strategy during training ($k=1$).}
\end{table*}

\begin{table*}[t]
  \small
  \centering
\begin{tabular}{clcccccccccc}
  \toprule
  \multicolumn{2}{c}{\multirow{2}{*}{\text{\ours}$_\text{xl}$}}& \multicolumn{5}{c}{Top-$k$} & & \multicolumn{4}{c}{Top-$p$}\\ %
  \cmidrule{3-7}\cmidrule{9-12}
  &              & $1^*$& $2$    & $4$    & $6$    & $8$    && $0.5\%$   & $1\%$    & $1.5\%$   & $2\%$  \\
  \midrule
  \multirow{2}{*}{A.R.} &\arguana       & 32.4 & 28.3 & 22.5 & 18.9 & 16.3 && 35.3 & 44.9 & 47.2 & 47.0 \\ %
  &\touche                              & 36.2 & 46.4 & 53.2 & 51.0 & 50.5 && 33.0 & 26.1 & 21.9 & 18.2 \\ %
  \cmidrule(lr){1-2}
  \multirow{3}{*}{F.C.} &\fever         & 72.9 & 75.6 & 70.9 & 63.2 & 56.0 && 72.9 & 56.9 & 48.2 & 34.1 \\ %
  &\climatefever                        & 18.7 & 21.2 & 23.2 & 23.9 & 24.2 && 18.8 & 15.2 & 16.6 & 14.3 \\ %
  &\scifact                             & 71.5 & 70.8 & 69.7 & 67.6 & 65.6 && 73.0 & 71.6 & 69.9 & 69.5 \\ %
  \cmidrule(lr){1-2}
  \multirow{5}{*}{Q.A.} &\nq            & 58.8 & 54.9 & 46.0 & 39.5 & 34.1 && 58.8 & 55.5 & 50.1 & 43.8 \\ %
  &\hotpotqa                            & 63.9 & 62.6 & 45.5 & 32.9 & 25.0 && 63.9 & 62.5 & 57.5 & 51.4 \\ %
  &\fiqa                                & 40.8 & 37.4 & 32.2 & 28.9 & 25.2 && 40.8 & 35.0 & 31.2 & 25.7 \\ %
  &\bioasq                              & 53.6 & 50.4 & 43.0 & 36.6 & 32.2 && 50.2 & 39.7 & 34.4 & 28.9 \\ %
  &\nfcorpus                            & 35.4 & 34.3 & 33.0 & 31.5 & 29.7 && 35.0 & 33.1 & 31.6 & 30.0 \\ %
  \cmidrule(lr){1-2}
  \multirow{4}{*}{MISC.} &\treccovid     & 75.1 & 80.6 & 80.5 & 78.1 & 73.5 && 72.5 & 69.4 & 62.0 & 54.0 \\ %
  &\scidocs       & 15.4 & 16.0 & 15.5 & 14.3 & 13.3 && 15.6 & 16.2 & 16.5 & 16.4 \\ %
  &\dbpedia      & 43.6 & 42.2 & 36.2 & 30.6 & 26.5 && 43.6 & 43.6 & 42.4 & 39.4 \\ %
  &\quora                               & 85.3 & 76.5 & 50.1 & 26.9 & 13.1 && 85.3 & 85.3 & 85.3 & 85.3 \\ %
  \midrule
  & Average                                & 50.3     & 49.8 & 44.4 & 38.9 & 34.7 && 49.9 & 46.8 & 43.9 & 39.8  \\ %
  
\bottomrule
\end{tabular}
\caption{\label{full_beir_xl}nDCG@10 on the BEIR benchmark with different $k$ and $p$ in \text{\ours}$_\text{xl}$. $^*$: alignment strategy during training ($k=1$).}
\end{table*}

\begin{table*}[t]
  \small
  \centering
\begin{tabular}{clcccccccccc}
  \toprule
  \multicolumn{2}{c}{\multirow{2}{*}{\text{\ours}$_\text{xxl}$}}& \multicolumn{5}{c}{Top-$k$} & & \multicolumn{4}{c}{Top-$p$}\\ %
  \cmidrule{3-7}\cmidrule{9-12}
  &              & $1^*$& $2$    & $4$    & $6$    & $8$    && $0.5\%$   & $1\%$    & $1.5\%$   & $2\%$  \\
  \midrule
  \multirow{2}{*}{A.R.} &\arguana       & 33.8 & 29.6 & 24.1 & 20.4 & 18.2 && 36.6 & 46.9 & 49.8 & 49.7 \\ %
  &\touche                              & 34.5 & 47.4 & 49.8 & 51.1 & 50.6 && 30.1 & 21.8 & 16.3 & 11.7 \\ %
  \cmidrule(lr){1-2}
  \multirow{3}{*}{F.C.} &\fever         & 74.2  & 77.0  & 73.9  & 67.9  & 62.2  && 75.2 & 47.1  & 36.2  & 24.1 \\ %
  &\climatefever                        & 19.7  & 21.6  & 23.7  & 23.2  & 24.3  && 19.7 & 12.0  & 12.3  & 9.3   \\
  &\scifact                             & 73.1 & 71.2 & 71.3 & 69.1 & 67.0 && 74.4 & 71.5 & 69.9 & 69.5 \\ %
  \cmidrule(lr){1-2}
  \multirow{5}{*}{Q.A.} &\nq            & 60.5  & 56.0  & 49.1  & 44.0  & 40.1  && 60.4  & 54.5  & 45.8  & 37.9  \\
  &\hotpotqa                            & 65.2  & 63.4  & 48.8  & 37.7  & 30.1  && 65.2  & 63.0  & 56.0  & 48.1  \\
  &\fiqa                                & 43.5 & 40.3 & 36.8 & 33.8 & 31.4 && 43.5 & 35.9 & 30.4 & 24.0 \\ %
  &\bioasq                              & 54.8  & 51.1  & 43.3  & 38.2  & 34.4  && 49.6  & 35.8  & 29.4  & 24.9  \\
  &\nfcorpus                            & 35.2 & 34.0 & 32.6 & 31.5 & 30.3 && 34.1 & 29.7 & 28.0 & 27.1 \\ %
  \cmidrule(lr){1-2}
  \multirow{4}{*}{MISC.} &\treccovid     & 75.8 & 81.4 & 80.1 & 75.6 & 71.9 && 75.1 & 65.4 & 54.0 & 45.9 \\ %
  &\scidocs       & 17.1 & 16.8 & 16.3 & 15.6 & 14.2 && 17.1 & 17.0 & 16.6 & 15.9 \\ %
  &\dbpedia      & 45.0 & 43.2 & 38.3 & 33.9 & 30.6 && 45.0 & 44.9 & 42.6 & 35.7 \\ %
  &\quora                               & 86.0  & 79.0  & 58.6  & 38.2  & 22.8  && 86.0 & 86.0  & 86.0  & 86.0 \\ %
  \midrule
  & Average                             & 51.3  & 50.9  & 46.2  & 41.4  & 37.7  && 50.8 & 45.1  & 40.9  & 36.4 \\ %
  
\bottomrule
\end{tabular}
\caption{\label{full_beir_xxl}nDCG@10 on the BEIR benchmark with different $k$ and $p$ in \text{\ours}$_\text{xxl}$. $^*$: alignment strategy during training ($k=1$).}
\end{table*}

\begin{table*}[t]
  \small
  \centering
\begin{tabular}{clcccccccccc}
  \toprule
  \multicolumn{2}{c}{\multirow{2}{*}{\text{\ours}$_\text{base}$}}& \multicolumn{5}{c}{Top-$k$} & & \multicolumn{4}{c}{Top-$p$}\\ %
  \cmidrule{3-7}\cmidrule{9-12}
  && $1$& $2$    & $4^*$    & $6$    & $8$    && $0.5\%$   & $1\%$    & $1.5\%$   & $2\%$  \\
  \midrule
  \multirow{2}{*}{A.R.} &\arguana       & 26.5 & 23.6 & 18.8 & 15.9 & 13.6 && 30.7 & 42.8 & 45.7 & 46.2 \\
  &\touche                              & 25.2 & 32.4 & 43.9 & 48.2 & 49.5 && 19.7 & 14.6 & 11.9 & 9.5  \\
  \cmidrule(lr){1-2}
  \multirow{3}{*}{F.C.} &\fever         & 59.0 & 67.0 & 72.8 & 73.6 & 73.0 && 59.0 & 34.3 & 25.4 & 18.4 \\
  &\climatefever                        & 14.7 & 17.5 & 20.9 & 22.8 & 23.7 && 14.7 & 7.8  & 7.1  & 5.1  \\
  &\scifact                             & 70.1 & 70.6 & 70.4 & 69.2 & 68.0 && 69.6 & 66.7 & 65.6 & 64.9 \\
  \cmidrule(lr){1-2}
  \multirow{5}{*}{Q.A.} &\nq            & 42.3 & 49.4 & 52.4 & 51.2 & 49.1 && 42.3 & 37.3 & 30.9 & 25.3 \\
  &\hotpotqa                            & 57.6 & 61.0 & 60.4 & 57.4 & 53.0 && 57.6 & 55.3 & 49.4 & 43.4 \\
  &\fiqa                                & 29.8 & 32.2 & 32.0 & 30.1 & 27.8 && 29.9 & 22.2 & 17.7 & 14.0 \\
  &\bioasq                              & 48.3 & 50.9 & 49.0 & 48.0& 45.9 && 41.1 & 26.4 & 21.1 & 17.4 \\
  &\nfcorpus                            & 31.8 & 33.5 & 33.9 & 33.6 & 33.0 && 29.9 & 26.3 & 25.7 & 25.2 \\
  \cmidrule(lr){1-2}
  \multirow{4}{*}{MISC.} &\treccovid    & 54.8 & 63.5 & 72.9 & 75.4 & 74.5 && 53.1 & 48.7 & 42.5 & 38.1 \\
  &\scidocs                             & 13.2 & 14.1 & 14.7 & 14.6 & 14.0 && 13.2 & 12.2 & 11.7 & 11.5 \\
  &\dbpedia                             & 31.5 & 39.1 & 42.1 & 39.9 & 36.9 && 31.4 & 31.4 & 29.7 & 25.1 \\
  &\quora                               & 82.7 & 80.5 & 72.1 & 57.4 & 41.9 && 82.7 & 82.7 & 82.7 & 82.7 \\
  \midrule
  & Average                             & 42.0 & 45.4 & 46.9 & 45.5 & 43.1 & & 41.1 & 36.3 & 33.4  & 30.5  \\
\bottomrule
\end{tabular}
\caption{nDCG@10 on the BEIR benchmark with different $k$ and $p$ in \text{\ours}$_\text{base}$. $^*$: alignment strategy during training ($k=4$).
}
\end{table*}

\begin{table*}[t]
  \small
  \centering
\begin{tabular}{clcccccccccc}
  \toprule
  \multicolumn{2}{c}{\multirow{2}{*}{\text{\ours}$_\text{base}$}}& \multicolumn{5}{c}{Top-$k$} & & \multicolumn{4}{c}{Top-$p$}\\ %
  \cmidrule{3-7}\cmidrule{9-12}
  && $1$& $2$    & $4$    & $6$    & $8$    && $0.5\%$   & $1\%^*$    & $1.5\%$   & $2\%$  \\
  \midrule
  \multirow{2}{*}{A.R.} &\arguana       & 28.4 & 23.7 & 17.9 & 14.2 & 11.1 && 32.9 & 45.3 & 48.0 & 47.7 \\ %
  &\touche                              & 38.1 & 51.3 & 53.0 & 51.3 & 50.0 && 35.0 & 26.6 & 22.6 & 17.1 \\ %
  \cmidrule(lr){1-2}
  \multirow{3}{*}{F.C.} &\fever         & 72.1 & 74.6 & 67.7 & 57.5 & 49.2 && 72.2 & 56.1 & 46.2 & 31.2 \\ %
  &\climatefever                        & 17.8 & 20.3 & 22.0 & 22.1 & 21.8 && 17.8 & 13.5 & 14.4 & 11.3 \\ %
  &\scifact                             & 69.0 & 67.9 & 64.8 & 61.5 & 57.2 && 70.5 & 70.3 & 68.1 & 63.9 \\ %
  \cmidrule(lr){1-2}

  \multirow{5}{*}{Q.A.} &\nq            & 52.6 & 48.2 & 36.4 & 26.4 & 20.2 && 52.5 & 50.4 & 44.7 & 38.2 \\ %
  &\hotpotqa                            & 60.6 & 57.7 & 35.1 & 21.1 & 13.1 && 60.7 & 59.6 & 53.9 & 47.6 \\ %
  &\fiqa                                & 33.7 & 29.4 & 23.5 & 18.8 & 15.8 && 33.6 & 29.8 & 26.2 & 20.6 \\ %
  &\bioasq                              & 40.3 & 48.4 & 39.9 & 33.1 & 26.4 && 50.0 & 40.3 & 35.3 & 28.8 \\ %
  &\nfcorpus                            & 34.7 & 33.7 & 32.1 & 30.2 & 27.8 && 34.3 & 32.7 & 31.6 & 29.6 \\ %
  \cmidrule(lr){1-2}
  \multirow{4}{*}{\rotatebox[origin=c]{0}{MISC.}} 
    &\treccovid     & 67.6 & 73.4 & 73.1 & 67.7 & 62.5 && 67.2 & 65.7 & 60.6 & 52.0 \\ %
  &\scidocs       & 13.9 & 14.2 & 13.1 & 11.5 & 9.9  && 14.1 & 14.6 & 14.8 & 14.9 \\ %
  &\dbpedia      & 41.3 & 40.3 & 29.5 & 21.2 & 14.0 && 41.2 & 41.3 & 40.3 & 37.1 \\ %
  &\quora                               & 83.6 & 60.8 & 22.8 & 8.9  & 4.3  && 83.6 & 83.6 & 83.6 & 83.6 \\ %
    \midrule

  & Average                                & 43.6 & 42.9 & 35.4 & 29.7 & 25.5 && 44.4 & 42.0 & 39.3 & 34.9 \\ %
  
\bottomrule
\end{tabular}
\caption{nDCG@10 on the BEIR benchmark with different $k$ and $p$ in \text{\ours}$_\text{base}$. $^*$: alignment strategy during training ($p=1\%$).}
\end{table*}

\begin{table*}[t]
  \small
  \centering
  \resizebox{\linewidth}{!}{%
\begin{tabular}{clccccccccccc}
  \toprule
  \multicolumn{2}{c}{\multirow{2}{*}{\text{\ours}$_\text{base}$}}& \multicolumn{5}{c}{$k$} & & \multicolumn{5}{c}{$\epsilon$}\\ %
  \cmidrule{3-7}\cmidrule{9-13}
  && $1^*$& $2$    & $4$    & $6$    & $8$    && $0.01$ & $0.02$   & $0.04$    & $0.06$   & $0.1$  \\
  \midrule
  \multirow{2}{*}{A.R.} &\arguana       & 30.8 & 26.3 & 20.1 & 15.8 & 12.4 && 30.8     & 33.2     & 38.8     & 43.5     & 50.1     \\ %
  &\touche                              & 37.4 & 50.1 & 52.0 & 51.1 & 49.0 && 37.4     & 35.0     & 29.8     & 26.9     & 19.4     \\ %
  \cmidrule(lr){1-2}
  \multirow{3}{*}{F.C.} &\fever         & 68.8 & 71.4 & 63.7 & 46.1 & 44.3 && 68.8     & 68.0     & 66.4     & 64.5     & 56.2     \\ %
  &\climatefever                        & 16.7 & 19.4 & 21.2 & 21.6 & 21.3 && 16.7     & 16.2     & 15.6     & 15.6     & 15.4     \\ %
  &\scifact                             & 69.8 & 68.2 & 64.7 & 61.8 & 57.0 && 69.8     & 69.5     & 70.3     & 71.0     & 70.1     \\ %
  \cmidrule(lr){1-2}

  \multirow{5}{*}{Q.A.} &\nq            & 52.0 & 47.4 & 35.0 & 25.8 & 18.5 && 52.0     & 51.6     & 50.8     & 50.0     & 46.0     \\ %
  &\hotpotqa                            & 59.6 & 56.1 & 33.4 & 20.6 & 12.6 && 59.6     & 59.5     & 60.0     & 61.0     & 59.8     \\ %
  &\fiqa                                & 32.9 & 29.6 & 23.0 & 18.6 & 14.5 && 32.9     & 32.7     & 32.9     & 32.4     & 30.9     \\ %
  &\bioasq                              & 50.4 & 47.2 & 38.0 & 31.4 & 24.5 && 50.4     & 50.4     & 50.8     & 50.4     & 43.7     \\ %
  &\nfcorpus                            & 34.1 & 33.6 & 31.6 & 29.6 & 27.4 && 34.1     & 34.4     & 34.2     & 34.2     & 33.5     \\ %
  \cmidrule(lr){1-2}
  \multirow{4}{*}{MISC.} 
 &\treccovid     & 63.9 & 73.2 & 72.5 & 67.7 & 62.8 && 63.9     & 65.5     & 63.6     & 63.8     & 54.2     \\ %
 &\scidocs       & 14.0 & 14.4 & 13.4 & 11.6 & 9.6  && 14.0     & 14.1     & 14.2     & 14.5     & 15.4     \\ %
  &\dbpedia      & 40.9 & 38.7 & 28.7 & 19.9 & 13.2 && 40.9     & 40.5     & 39.9     & 39.7     & 38.6     \\ %
  &\quora                               & 82.8 & 61.2 & 21.5 & 7.4  & 3.5  && 82.8     & 83.1     & 83.5     & 84.0     & 85.0     \\ %
  \midrule
  & Average                                & 43.6 & 42.4 & 34.6 & 28.6 & 24.7 && 43.6 & 43.6 & 43.4 & 43.4 & 41.2 \\ %
  
\bottomrule
\end{tabular}
}
\caption{nDCG@10 on the BEIR benchmark with \text{\ours}$_\text{base}$ and differentiable alignment (\Cref{sec:diffalign}). $^*$: alignment strategy during training ($k=1$). 
}
\end{table*}

\end{document}